\documentclass{article}
\usepackage[numbers,square,comma]{natbib}
\usepackage[preprint]{neurips_2026}
\usepackage[utf8]{inputenc}
\usepackage[T1]{fontenc}
\usepackage{hyperref}
\usepackage{url}
\usepackage{booktabs}  
\usepackage{amsfonts}  
\usepackage{nicefrac}  
\usepackage{microtype} 
\usepackage{xcolor}
\usepackage{colortbl}
\usepackage{graphicx}  
\usepackage{amsmath}
\usepackage{float}
\usepackage{wrapfig} 
\usepackage{subcaption}
\setcitestyle{numbers}
\usepackage[ruled,vlined]{algorithm2e}
\usepackage{listings}
\usepackage{multirow}
\usepackage{adjustbox}
\usepackage{comment}
\title{Do Speech BCIs Need Larger Models? Rethinking Neural Decoding beyond Scaling}

\author{
\textbf{Zehui Feng}\textsuperscript{1,2}
\quad
\textbf{Weichuan Wang}\textsuperscript{1}
\quad
\textbf{Xiaohan Chen}\textsuperscript{1}
\quad
\textbf{Cuntai Guan}\textsuperscript{2,*}
\quad
\textbf{Ting Han}\textsuperscript{1,3,*}
\\[1mm]
\textsuperscript{1}Shanghai Jiao Tong University
\\
\textsuperscript{2}Nanyang Technological University
\\
\textsuperscript{3}Zhejiang University
}

\begin{document}

\maketitle
\begin{center}
\includegraphics[width=0.9\linewidth]{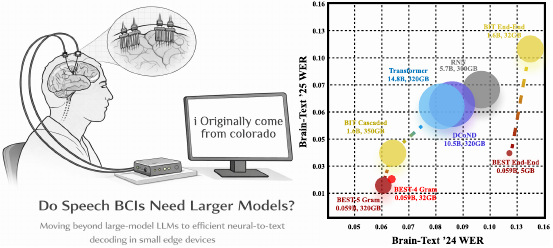}

\vspace{1mm}
\small{\textbf{Figure 1.} Method Parameter (B), RAM (GiB), and Word Error Rate (WER) on Brain-to-Text '24 and '25. Smaller circles indicate lower resource consumption.}
\end{center}
\setcounter{figure}{1}
\begin{abstract}
Recent speech brain-computer interfaces (BCIs) increasingly rely on large models and complex multi-stage pipelines for neural decoding. However, in speech BCI, where training data are scarce, noisy, and expensive to collect, the benefits of model scaling remain unclear. This work revisits a fundamental question: \textbf{are large models truly necessary for accurate neural decoding?}
We propose \textbf{\textit{BEST}} (\textbf{\textit{B}}rain-\textbf{\textit{E}}ncoding-\textbf{\textit{S}}peech-\textbf{\textit{T}}ext), a two-stage framework for efficient neural-to-text decoding. Instead of relying on large generative models or phoneme-centric pipelines, \textbf{\textit{BEST}} introduces hierarchical intermediate representations and a compact speech-oriented decoder that directly maps neural activity to text. This design preserves essential structure while significantly reducing computational cost, enabling robust generalization in low-resource settings and practical deployment under constrained resources.
On the Brain-to-Text ’24 and ’25 benchmarks, \textbf{\textit{BEST}} achieves word error rates (WERs) of 6.19\% and 1.62\%, respectively, using only 3.85\% of the parameters required by comparable methods. Beyond these results, we provide three key insights: (1) decoder scaling yields non-monotonic gains; (2) cross-modal alignment is more critical than model size; and (3) overfitting degards sequence-level performance. Representation analysis further reveals a consistent neural→audio→text alignment pathway.
\end{abstract}
\section{Introduction}
Speech brain-computer interfaces (BCIs) offer a promising pathway for restoring communication in individuals with severe motor impairments \cite{Wairagkar2025, Singh2025}. By directly mapping neural activity to language, these systems bypass damaged neuromuscular pathways and enable brain-to-text communication \cite{Ascoli2025}. Recent advances in neural decoding\cite{yin2026brainvistamodelingnaturalisticbrain} have substantially improved transcription quality, bringing speech BCIs closer to practical deployment \cite{zhang2026, feghhi, li2024}.
With the rapid advancement of large language models (LLMs)\cite{zhang2026diffusion, jiang2026robustmultimodalphysiologicalfoundation,lin2026vcast,10.1007/978-3-032-37232-1_16}, most existing approaches adopt a multi-stage pipeline: neural signals are first decoded into phoneme sequences via supervised classification \cite{li2024}, and then converted into text using LLMs, often with additional rescoring \cite{openai2024gpt4technicalreport, zhang2022optopenpretrainedtransformer}. While effective due to the increasingly strong linguistic priors provided by rapidly evolving LLMs, this paradigm suffers from two fundamental limitations.

First, it is poorly aligned with real-world deployment constraints. Practical BCI systems require low-latency and resource-efficient inference \cite{Lzw2025, Y1126}, yet reliance on large pretrained models and memory-intensive rescoring introduces substantial computational overhead, limiting feasibility on edge or clinical hardware. Second, the pipeline imposes a discrete phoneme bottleneck that discards rich neural information. In data-scarce and noisy BCI settings, collapsing high-dimensional neural signals into hard phoneme labels amplifies overfitting and suppresses fine-grained linguistic structure \cite{feghhi}. Moreover, the strict separation between intermediate prediction and sentence generation prevents effective use of multi-level neural-linguistic representations during decoding.

These challenges point to a central requirement: an effective speech BCI must achieve high accuracy while remaining lightweight, data-efficient, and robust under limited supervision. To this end, we propose \textbf{\textit{BEST}} (\textbf{\textit{B}}rain-\textbf{\textit{E}}ncoding-\textbf{\textit{S}}peech-\textbf{\textit{T}}ext), a two-stage framework that replaces stage-wise symbolic pipelines with hierarchical feature-level integration. In Stage 1, multi-level neural-linguistic representations are fused through a lightweight integration module, allowing phonological and lexical information to interact before discrete decoding and mitigating overfitting induced by phoneme-level supervision. In Stage 2, the fused representations are translated by a compact speech-oriented decoder, followed by beam search with a lightweight n-gram language model for efficient hypothesis scoring. This design preserves informative neural structure while introducing linguistic regularization without the computational burden of large-scale reranking.

We evaluate BEST on the Brain-to-Text ’24 and ’25 benchmarks, achieving word error rates (WERs) of 6.19\% and 1.62\%, respectively. Despite using only 3.85\% of the parameters of comparable methods, BEST consistently improves decoding performance while remaining suitable for edge-class deployment. Representation analyses further reveal a coherent neural → linguistic → text alignment process underlying the model’s predictions.

In summary, our contributions are threefold:

1. We propose BEST, a lightweight and practical framework for speech BCI that achieves strong performance with only 3.7\% of the parameters of existing approaches.

2. We identify the discrete phoneme bottleneck and introduce a hierarchical feature-level decoding paradigm that preserves and integrates multi-level neural-linguistic representations.

3. We systematically analyze the role of decoder scale in speech BCI, showing that performance gains depend more on structured alignment than on model size under realistic constraints.
\section{Related Work}

\textbf{Speech brain-computer interfaces.}
Speech BCIs decode neural activity from auditory and motor cortices into text. Early systems use RNN-based models to map neural signals to intermediate units (e.g., phonemes or characters), followed by statistical language models for text generation \cite{CHEN1999359, Bouchard2013, Willett2023, NEURIPS2023_83a14a36, Littlejohn2025, KUNZ20254658}. Later work incorporates context-dependent modeling (e.g., diphones) to capture local phonetic transitions \cite{li2024}. More recent approaches adopt transformer architectures for long-range dependency modeling and streaming or end-to-end decoding, often combined with LLMs \cite{feghhi, Littlejohn2025, Feng_2024}. Despite these advances, most methods rely on discrete intermediate representations and fail to preserve fine-grained temporal and spectral structure in neural signals \cite{doi10}.
\begin{figure*}[h]
    \centering
    \includegraphics[width=1\linewidth]{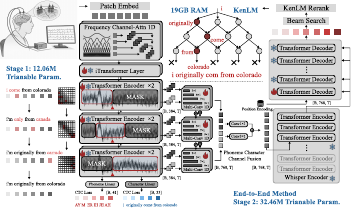}
    \vspace{-2mm}
    \caption{\textbf{\textit{Overall illustration of \textit{BEST}}. The framework consists of four core modules: (1) \textit{Hierarchical Transformer Encoder} (Stage 1): it employs frequency-channel attention and iTransformer layers to encode neural activity into phoneme/character representations with hierarchical temporal masking; (2) \textit{Phoneme-Character Fusion Block} (Stage 1): a lightweight convolutional fusion module aggregates hierarchical neural representation to bridge neural and linguistic spaces; (3) \textit{Lightweight Speech Translator} (Stage 2): a lightweight speech translator generates autoregressive text sequences from fused embeddings; and (4) \textit{KenLM Reranking} (Stage 2): a custom 4-gram language model resolves ambiguities from noisy neural signals for final selection.}}
    \label{fig:1}
\end{figure*}

\textbf{Large language models for BCI.}
LLMs have recently enabled neural-to-text decoding, semantic reconstruction, and cross-modal alignment\cite{ICLR2026_b16e55ac,11516011,10.1007/978-3-032-38085-2_12}. Prior work explores LLM fine-tuning for neural decoding \cite{mishra, tang2026eegdlitedatasetdistillationefficient, ding2025brainprolargescalebrainstateaware, puah2025eegdmeegrepresentationlearning} and autoregressive reconstruction from fMRI signals \cite{Ye2025}. LLMs have also been used to align neural and linguistic representations \cite{liu2025}, model semantic organization \cite{lu2025, feng2025}, and support applications such as spellers and rehabilitation systems \cite{Andrea, jin2025, hossain2025,li2026frozen,li2026memoryathena}. However, current approaches face key limitations: (1) end-to-end LLM decoding ignores structural mismatch between neural signals and linguistic embeddings; (2) large-scale inference incurs high computational and memory cost \cite{zhang2026,shen2026cortexvideo}; and (3) weakly constrained decoding reduces robustness under noisy neural inputs.

\textbf{Real-world BCI deployment.}
Practical BCI systems require efficient, robust, and accurate decoding under resource constraints \cite{Lzw2025, Y1126,lan2026braintracetracinglongitudinalmultimodal,yu2026evotreenadgenealogyguidedevolutionllmdriven}. Existing lightweight models mainly target motor or intention decoding and do not generalize well to speech \cite{liu2024, 10452347}. Classical methods (e.g., random forests) are efficient but cannot model the sequential dependencies required for speech \cite{lin2025towards,FENG2025103363}. Lightweight neural models face a core trade-off: limited capacity weakens representation learning, while increased complexity leads to overfitting and unstable optimization in low-resource settings \cite{angrick2025realtime,10904227}. This highlights the need for architectures that balance efficiency with structurally aligned neural-linguistic modeling.
\section{Method}
In this section, we present a lightweight speech BCI framework with a hybrid decoding strategy. The overall architecture comprises a hierarchical transformer-based neural encoder, a lightweight phoneme-character fusion block, a speech translator, and a custom language model (Figure \ref{fig:1}). The neural encoder is pretrained to map neural signals into both phoneme-level and character-level representations, providing robust linguistic features. At inference time, encoder representations are fed into a Whisper-based decoder to generate an n-best list of hypotheses, which are subsequently rescored with a 4-gram KenLM for final selection.
\subsection{Architecture}
\textbf{Stage 1: hierarchical transformer encoder.}
We introduce a hierarchical Transformer encoder \cite{transformer} tailored for neural speech decoding, combining frequency-aware channel modulation with structured temporal attention. Neural signals are partitioned into temporal patches and embedded as token sequences. To incorporate spectral structure, we apply a multi-frequency channel attention mechanism (Appendix \ref{sec:multi-frequency}) that projects inputs onto predefined frequency bases, aggregates channel-wise statistics, and produces adaptive reweighting coefficients, enabling frequency-selective feature refinement.

The resulting embeddings are first processed by an iTransformer \cite{itransformer} to capture global dependencies and stabilize latent representations, followed by a hierarchical Transformer with structured attention (Appendix \ref{sec:hierarchical}). Temporal dependencies are regulated through three complementary masking schemes: past-oriented masking for historical context, local bidirectional masking for short-range interactions, and future-oriented masking to incorporate forward cues. These masks are applied sequentially to progressively refine the temporal receptive field, enabling the model to capture localized temporal patterns while maintaining global consistency. The final representations are projected to both phoneme-level and character-level outputs via CTC heads, providing complementary supervision at different linguistic granularities.

\textbf{Stage 2: phoneme–character fusion.}
To integrate multi-level linguistic representations, we design a lightweight multi-scale convolutional fusion module that aggregates features across encoder depths. Phoneme and character streams are processed through parallel convolutional branches with different receptive fields, capturing complementary temporal patterns. Within each stream, residual aggregation enables iterative refinement and cross-level information flow while preserving temporal structure. The fused features are then concatenated and projected into a unified representation through lightweight convolutions.

Compared to additional attention layers, this design is more computationally efficient and better suited to low-resource, high-noise BCI settings. Multi-scale convolutions capture both local and longer-range dependencies without the cost of full self-attention. The fusion module also serves as a projection interface, aligning neural representations with the acoustic embedding space and facilitating downstream linguistic decoding.

\textbf{Stage 2: lightweight speech translator.}
To enable efficient neural-to-text mapping without large language models, we adopt a compact speech translation module based on Whisper \cite{whisper}. The fused representations are projected into the acoustic embedding space and processed by the upper layers of the Whisper encoder, aligning them with pretrained speech representations and introducing implicit acoustic priors without requiring explicit audio input. The transformed embeddings are then decoded autoregressively by the Whisper decoder.

The model is trained using next-token prediction with teacher forcing. During inference, text is generated conditioned on neural-derived embeddings, optionally with beam search to improve sequence quality. This design replaces LLM-based decoding with a pretrained speech model, enabling direct neural-to-text transcription with substantially lower computational cost.

\textbf{Stage 2: lightweight language model reranking.} Neural-derived embeddings are inherently noisy, leading to ambiguous candidate sequences. To improve robustness, we apply lightweight n-gram rescoring using a custom 4-gram KenLM \cite{kenLM}. Compared to standard large-scale n-gram models (e.g., 5-gram models), this approach provides effective linguistic regularization with minimal overhead. It improves disambiguation and transcription accuracy while preserving real-time feasibility for BCI deployment.

\vspace{-2mm}
\subsection{Training objective}
\textbf{Pretraining stage.}
In contrast to masked reconstruction approaches such as BIT \cite{zhang2026}, we do not employ masked autoencoding. While masking promotes contextual inference, neural speech decoding depends more on preserving temporal–spectral structure and higher-level phonetic and linguistic information (Appendix \ref{sec:masked}). To improve generalization, we adopt extensive data augmentation and hierarchical pretraining at both phoneme and word levels, enabling the encoder to learn compact representations that capture both fine-grained acoustic structure and higher-level linguistic information from neural signals.

\textbf{Fine-tuning stage.}
For sentence-level decoding, we freeze the neural encoder and Whisper encoder to preserve stable representations. Phoneme- and character-level features are injected into the upper layers of the Whisper encoder, while only the first layer of the Whisper decoder and the fusion module are updated. This targeted optimization restricts learning to lightweight components, reducing overfitting and training instability while enabling effective adaptation to neural-derived representations for accurate transcription.
\section{Evaluation}
\subsection{Dataset}
We evaluate our method on the Brain-to-Text ’24 \cite{Willett2023} and ’25 \cite{Card2024Accurate} benchmarks for attempted speech decoding. The ’24 dataset contains 12,100 sentences from participant T12 collected over 25 sessions using intracortical microelectrode arrays implanted in speech-related regions (Appendix \ref{sec:Data preprocessing}). The ’25 dataset includes 10,948 sentences from participant T15 across 45 sessions with 256 electrodes in the speech motor cortex. In both datasets, participants were visually prompted with sentences and asked to attempt speech following a cue, with decoding performed on the corresponding neural activity.

The ’24 dataset consists of conversational sentences with attempted vocalized speech at approximately 62 words per minute. In contrast, the ’25 dataset spans more diverse corpora, and custom sets, and includes both attempted vocalized and silent speech. Neural features include threshold crossing spike counts and spike-band power, extracted at 20 ms resolution and normalized across sessions to reduce non-stationarity.
\vspace{-2mm}
\subsection{Baselines}
We benchmark our method against 4 state-of-the-art baselines in neural speech BCIs:

\textbf{1.} \textit{\textbf{RNN Baseline}} \cite{Willett2023} employs a 5-layer bidirectional GRU to map neural activity to phonemes. It generates text via a 5-gram language model with OPT rescoring.\\
\textbf{2.} \textit{\textbf{Transformer Baseline}} \cite{feghhi} adopts a causal transformer enhanced with time-masking augmentation to improve generalization. It operates on non-overlapping neural patches with relative positional encoding and supports real-time streaming inference.\\
\textbf{3.} \textit{\textbf{DCoND}} \cite{li2024} improves phoneme decoding by modeling diphone contexts rather than individual phonemes. It uses a GRU encoder to estimate diphone distributions and leverages an ensemble of decoders with an LLM for correction.\\
\textbf{4.} \textit{\textbf{BIT}} \cite{zhang2026} is an end-to-end speech decoding framework built on a cross-species, cross-task pretrained transformer. The encoder is first self-supervised pretrained via masked temporal patch reconstruction, then fine-tuned for phoneme decoding with CTC loss.
\subsection{Results}
\textbf{Performance comparison on speech BCI decoding:} We evaluate \textit{BEST} on the Brain-to-Text ’24 and ’25 benchmarks (Table \ref{tab:sota_comparison}), reporting model size, memory consumption, phoneme error rate (PER), character error rate (CER), and WER on both validation and held-out test sets. Following prior work \cite{zhang2026}, we consider both end-to-end decoding and cascaded LM pipelines. \textit{BEST} is instantiated in three configurations reflecting different accuracy-efficiency trade-offs:

\textbf{1.} \textit{\textbf{Lightweight \textit{BEST}}} is an end-to-end pipeline without external language models that minimizes computational cost while preserving robutsness;\\
\textbf{2.} \textit{\textbf{Practical \textit{BEST}}} incorporates a 4-gram LM to balance linguistic regularization and efficiency;\\
\textbf{3.} \textit{\textbf{Robust \textit{BEST}}} employs a 5-gram LM to maximize accuracy under higher memory budgets.

The lightweight configuration consistently outperforms the end-to-end baseline of BIT despite using substantially fewer parameters and eliminating reliance on large pretrained language models, demonstrating the effectiveness of hierarchical feature-level decoding. Incorporating lightweight n-gram LMs further reduces WER, with the practical and robust variants achieving 2.92\% and 1.62\% WER on the ’25 benchmark, respectively. The results reveal that performance gains are primarily driven by improved neural representation learning rather than increased capacity to communicate with LMs. Notably, \textit{BEST} significantly narrows the gap between end-to-end and cascaded pipelines, indicating that mitigating the discrete phoneme bottleneck and preserving hierarchical neural-linguistic representations are critical for robust and efficient speech BCI decoding.

\textbf{LLM decoder scaling and adaptation analysis: }We systematically study the effect of decoder scale and adaptation strategy in end-to-end speech BCI (Table \ref{tab:llm_decoder}). Across all settings, we find that decoder scaling exhibits a distinctly non-monotonic trend. Small decoders such as Whisper-Tiny \cite{radford2022} and DistilGPT2 \cite{sanh2019distilbert} are capacity-limited, while larger text-based LLMs fail to deliver consistent gains with increasing scale. Instead, mid-scale decoders such as Qwen2-1.5B \cite{yang2024qwen2technicalreport} achieve the best performance, revealing a non-monotonic scaling trend. This suggests that, in speech BCI, increasing model capacity can amplify overfitting to noisy neural signals rather than improve generalization, making capacity-efficient decoders more effective than maximally scaled models.
\begin{table*}[!t]
  \centering
  \caption{\textbf{We compare the speech-BCI framework on the Brain-to-Text '24 and '25 hold-out set. Gray background indicates the pipeline without the language model. Some results lack detailed data as the corresponding code is unavailable; see the competition leaderboard for more details. $\star$, $\diamond$, $\bullet$ denote the most lightweight network, the most robust network, and the most practical network, respectively.}}
  \label{tab:sota_comparison}
  \large
  \resizebox{\linewidth}{!}{
  \begin{tabular}{l l| c |c |c |c |c |c |c |c}
    \toprule[1pt]
    \multirow{2}{*}{\textbf{Model}} & \multirow{2}{*}{\textbf{\#Param, RAM}} & \multicolumn{4}{c}{\textbf{Brain-to-text '24 T12}} & \multicolumn{4}{c}{\textbf{Brain-to-text '25 T15}} \\
    \cmidrule(lr){3-6} \cmidrule(lr){7-10}
    & & \textbf{PER$\downarrow$} & \textbf{CER$\downarrow$} & \textbf{Valid WER$\downarrow$} & \textbf{WER$\downarrow$} & \textbf{PER$\downarrow$} & \textbf{CER$\downarrow$} & \textbf{Valid WER$\downarrow$} & \textbf{WER$\downarrow$} \\
    \midrule
    RNN(Baseline)~\cite{Willett2023} & 5.70B, 300GiB & 0.2122 & 0.2561 & 0.2271 & 0.0976 & 0.1310 & 0.2075 & 0.1127 & 0.0667 \\
    Transformer \cite{feghhi} & 14.79B, 320GiB & 0.1628 & 0.1732 & 0.1121 & 0.0818 & 0.1081 & 0.1573 & 0.0934 & 0.0612 \\
    DCoND \cite{li2024} & 10.50B, 320GiB & 0.1534 & - & - & 0.0860 & - & - & - & - \\
    \rowcolor{gray!20} BIT \cite{zhang2026} & 1.53B, 32GiB & 0.1439 & - & 0.1607 & 0.1567 & 0.0712 & - & 0.0879 & 0.1106 \\
    BIT \cite{zhang2026} & 1.53B, 350GiB & - & - & - & 0.0635 & - & - & - & 0.0406 \\
    \midrule
    \rowcolor{gray!20} \textit{BEST}$\star$ & 0.059B, 5GiB & \textbf{0.1321} & \textbf{0.1436} & 0.1504 & 0.1261 & \textbf{0.0688} &\textbf{0.0896} & 0.0681 & 0.0517 \\
    \textit{BEST}+5-Gram LM$\diamond$ & 0.059B, 320GiB &\textbf{0.1321} & \textbf{0.1436}& \textbf{0.0721} & \textbf{0.0619} & \textbf{0.0688} &\textbf{0.0896} & \textbf{0.0257} & \textbf{0.0162} \\
    \textit{BEST}+4-Gram LM$\bullet$ & 0.059B, 32GiB &\textbf{0.1321} & \textbf{0.1436}& 0.0764 & 0.0654 & \textbf{0.0688} &\textbf{0.0896} & 0.0292 & 0.0196 \\
    \bottomrule[1pt]
  \end{tabular}
  }
\end{table*}
\begin{table*}[!t]
  \centering
  \caption{\textbf{Performance comparison across different decoders with parameter counts and fine-tuning strategies on T15. Gray
background indicates the audio-based LLM decoders.}}
  \label{tab:llm_decoder}
  \small
  \resizebox{\linewidth}{!}{
  \begin{tabular}{l|cc|c|c|c}
    \toprule[1pt]
    \multirow{2}{*}{\textbf{Decoder}} & \multirow{2}{*}{\textbf{\#Params}} & \multirow{2}{*}{\textbf{Dimension}} & \multicolumn{3}{c}{\textbf{Valid WER$\downarrow$}} \\
    \cmidrule(lr){4-6}
    & & & \textbf{Full fine-tuning} & \textbf{Partial fine-tuning} & \textbf{Frozen} \\
    \midrule
    \rowcolor{gray!20} Whisper-tiny \cite{radford2022}               & 34M    & 384   & 0.3678 & 0.2845 & 0.2996 \\
    \rowcolor{gray!20} Whisper-base \cite{radford2022}               & 74M    & 512   & 0.2409 & 0.2183 & 0.2235 \\
    DistilGPT2 \cite{sanh2019distilbert}                 & 82M    & 768   & 0.2091 & 0.1930 & 0.2006 \\
    \rowcolor{gray!20} Whisper-large \cite{radford2022}              & 1.55B  & 1280  & 0.1461 & 0.1460 & 0.1572 \\
    \rowcolor{gray!20} Whisper-medium   \cite{radford2022}           & 769M   & 1024  & 0.1351 & 0.1309 & 0.1366 \\ 
    Fine-tuned DistilGPT2 \cite{sanh2019distilbert}      & 82M    & 768   & 0.1323 & 0.1152 & 0.1363 \\
    Llama3.1 \cite{touvron2023llamaopenefficientfoundation} & 8B & 4096 & 0.1256 & 0.1163 & 0.1099 \\
    Fine-tuned Llama3.1 \cite{touvron2023llamaopenefficientfoundation} & 8B & 4096 & 0.1209 & 0.1131 & 0.1052 \\
    Qwen2 \cite{yang2024qwen2technicalreport}                 & 1.5B   & 1536  & 0.1133 & 0.1091 & 0.1050 \\
    Fine-tuned Qwen2 \cite{yang2024qwen2technicalreport}       & 1.5B   & 1536  & 0.1103 & 0.0967 & 0.0938 \\
    \rowcolor{gray!20} Aero1-Audio \cite{li2025aero}          & 1.5B   & 1280  & 0.1018 & 0.0989 & 0.0830 \\
    \rowcolor{gray!20} Whisper-small \cite{radford2022} (\textbf{\textit{BEST}})        & 244M   & 768   & \textbf{0.0764} & \textbf{0.0681} & \textbf{0.0795} \\
    \bottomrule[1pt]
  \end{tabular}
  }
\vspace{-3mm}
\end{table*}
A second key observation is that pretraining modality is decisive. Audio-oriented models (e.g., Aero1-Audio \cite{li2025aero}, Whisper \cite{radford2022}) consistently outperform text-only LLMs of similar or larger size, indicating that representation alignment with neural inputs, rather than model scale, dominates decoder performance. Even with additional projection layers for adaptation, text-only LLMs \cite{touvron2023llamaopenefficientfoundation,yang2024qwen2technicalreport} remain mismatched to the distribution of neural features. In contrast, speech-aligned pretraining produces representations inherently compatible with neural-derived embeddings.

We further observe a scale-dependent adaptation pattern. Full fine-tuning is consistently suboptimal, reflecting instability under data-limited and high-noise conditions. For compact decoders, partial fine-tuning is sufficient to align representations, whereas larger models favor fully frozen decoding, as pretrained priors are degraded by task-specific adaptation. Decoder dimensionality (768) has a limited effect on performance. The encoder dimension is selected via hyperparameter tuning (Appendix \ref{sec:Hyperparameter scanning on encoder dimension}), as increasing dimensionality exacerbates overfitting in the neural encoding stage.

\textbf{Encoder--decoder factorization.}
To disentangle the contributions of neural representation quality and downstream decoding, we factorize the encoder and decoder choices while keeping the remaining pipeline fixed (Table~\ref{tab:encoder_decoder_factorization}).
The BEST encoder consistently outperforms Transformer and GRU encoders under both Whisper-small and Aero, showing that its gain is not specific to a particular downstream decoder.
Conversely, Whisper-small consistently outperforms the substantially larger Aero decoder for every encoder, with WER gaps of 0.0308--0.0408.
Thus, increasing decoder capacity does not compensate for weak neural representations; both the quality of the neural encoder and its compatibility with the downstream speech decoder determine sequence-level performance.
Notably, the decoder gap decreases from 0.0408 with GRU to 0.0308 with BEST, suggesting that better-structured neural representations are also less sensitive to the choice of downstream decoder.
These results further support our central finding that representation quality and cross-stage compatibility, rather than decoder scale alone, dominate speech BCI decoding.

\textbf{Ablation analysis}: We observe a monotonic improvement as each component is introduced (Table \ref{tab:ablation}). Hierarchical attention improves temporal dependency modeling, while frequency attention further enhances spectral selectivity. The largest gain arises from the phoneme-character fusion module, showing that robust decoding depends on integrating multi-level linguistic representations rather than isolated phoneme prediction. \textit{BEST} also achieves lower validation CTC loss (Appendix \ref{sec:Hierarchical layer allocation} and \ref{sec:Validation CTC loss and CER analysis}), providing independent evidence of better-aligned neural representations and more stable optimization. The detailed architecture and hyperparameter settings are also provided (Appendix \ref{sec:Implementation details}, \ref{sec:RNN baseline hyperparameter}, \ref{sec:LLM decoder hyperparameter}, and \ref{sec:Hyperparameter}).

\begin{wraptable}{r}{8.3cm}
\setlength{\tabcolsep}{4pt}
\small
\centering
\vspace{-2mm}
\caption{\textbf{Encoder--decoder factorization on Brain-to-Text ’25 validation set.}}
\label{tab:encoder_decoder_factorization}
\vspace{-1mm}

\resizebox{\linewidth}{!}{
\begin{tabular}{lccc}
\specialrule{1pt}{0pt}{0pt}
\textbf{Neural encoder}
& \textbf{Whisper-small}
& \shortstack{\textbf{Audio-LLM}}
& \textbf{Decoder gap} \\
\hline
\textit{BEST} encoder
& \textbf{0.0681}
& \textbf{0.0989}
& \textbf{0.0308} \\
Transformer encoder
& 0.0862
& 0.1217
& 0.0355 \\
GRU encoder
& 0.1054
& 0.1462
& 0.0408 \\
\specialrule{1pt}{0pt}{0pt}
\end{tabular}
}
\vspace{2mm}
\caption{\textbf{Comparison of \textit{BEST}'s different components on Brain-to-Text ’24 and ’25.}}
\scalebox{0.92}{
\begin{tabular}{lcccc}
\specialrule{1pt}{0pt}{0pt}
\multirow{2}{*}{\textbf{Model}} & \multicolumn{2}{c}{\textbf{T12}} & \multicolumn{2}{c}{\textbf{T15}} \\
\cmidrule(lr){2-3} \cmidrule(lr){4-5}
& \textbf{PER$\downarrow$} & \textbf{CER$\downarrow$} & \textbf{PER$\downarrow$} & \textbf{CER$\downarrow$} \\
\hline
1. w. Transformer & 0.1628 & 0.1732 & 0.1081 & 0.1573 \\
2. w. Hierarical Attention Mask & 0.1538 & 0.1597 & 0.0835 & 0.1103 \\
3. w. Frequency Attention Block & 0.1321 & 0.1436 & 0.0688 & 0.0896 \\
4. w. Phoneme-Character Fusion & \textbf{0.1241} & \textbf{0.1394} & \textbf{0.0612} & \textbf{0.0758} \\
\specialrule{1pt}{0pt}{0pt}
\label{tab:ablation}
\end{tabular}
}
\caption{\textbf{Performance comparison of well-fitted and over-fitted settings on validate PER, CER, and WER.}}
\scalebox{0.92}{
\begin{tabular}{lcccccc}
\specialrule{1pt}{0pt}{0pt}
\multirow{2}{*}{\textbf{Model}} & \multicolumn{3}{c}{\textbf{’24 T12}} & \multicolumn{3}{c}{\textbf{’25 T15}} \\
\cmidrule(lr){2-4} \cmidrule(lr){5-7}
& \textbf{PER} & \textbf{CER} & \textbf{WER} & \textbf{PER} & \textbf{CER} & \textbf{WER} \\
\hline
Well-fitted & 0.1321 & 0.1436 & \textbf{0.1504} & 0.0688 & 0.0896 & \textbf{0.0681} \\
\rowcolor{gray!15} Over-fitted & \textbf{0.1219} & \textbf{0.1324} & \underline{0.1692} & \textbf{0.0621} & \textbf{0.0825} & \underline{0.0803} \\
\specialrule{1pt}{0pt}{0pt}
\label{tab:fit_comparison}
\end{tabular}
}

\caption{\textbf{Ablation performance on training strategy, loss setting, and Gram-LM settings.}}
\scalebox{0.85}{
\begin{tabular}{lcc}
\specialrule{1pt}{0pt}{0pt}
\textbf{Model Setting} & \textbf{T12 WER$\downarrow$} & \textbf{T15 WER$\downarrow$} \\
\hline
\rowcolor{gray!15} \textit{BEST}: End-to-End training & 0.3415 & 0.2071 \\
\rowcolor{gray!15} \textit{BEST}: End-to-End + Multi-task Loss & 0.2253 & 0.1067 \\
\rowcolor{gray!15} \textit{BEST}: End-to-End + Multi-task + 5-Gram & 0.2060 & 0.0895 \\
\textit{BEST}: Two-Stage training & 0.1504 & 0.0681 \\
\textit{BEST}: Two-Stage + 5-Gram LM & \textbf{0.0721} & \textbf{0.0257} \\
\textit{BEST}: Two-Stage + 4-Gram KenLM & 0.0764 & 0.0292 \\
\rowcolor{gray!15} Transformer: Two-Stage + 5-Gram & 0.0852 & 0.0381 \\
\rowcolor{gray!15} GRU: Two-Stage + 5-Gram & 0.1105 & 0.0541 \\
\specialrule{1pt}{0pt}{0pt}
\end{tabular}
}
\label{tab:training_ablation}
\end{wraptable}
\textbf{Overfitting analysis}: The overfitting model achieves lower PER and CER but degrades WER (Table \ref{tab:fit_comparison}), exposing a misalignment between local phoneme accuracy and sequence-level decoding. We attribute to representation collapse induced by overfitting. The visualization of output probability heatmap (Appendix \ref{sec:Phoneme probability visualization}) shows that the overfitting model exhibits highly concentrated and fragmented activations, reflecting overconfident and poorly calibrated predictions. These representations lack temporal coherence and fail to capture structured phoneme transitions, resulting in inconsistent sequence reconstruction. In contrast, the well-fitting model produces smoother and more distributed activations, preserving coherent neural-linguistic structure. This reveals a fundamental property of speech BCI: overfitting increases local discriminability while degrading global consistency. Reduced phoneme error does not translate to better decoding, as the resulting representations become less compatible with downstream inference. This explains why stronger phoneme supervision can harm end-to-end performance and supports the design of \textit{BEST} in preserving structured, multi-level representations rather than relying on discrete phoneme predictions.

\textbf{Training strategy and LM-controlled analysis.}
Two-stage training consistently outperforms end-to-end optimization on both T12 and T15 (Table~\ref{tab:training_ablation}), while multi-task supervision only partially closes the gap. This suggests that jointly learning neural representations and sequence generation is prone to overfitting under noisy, data-limited supervision. In contrast, two-stage training first stabilizes phoneme- and character-level representations and then optimizes sentence-level decoding on top of this structured latent space.

Language modeling provides additional sequence-level refinement, but does not explain the advantage of \textit{BEST}. Both 4-gram \cite{kenLM} and 5-gram \cite{Willett2023} LMs reduce WER, with the 5-gram LM yielding the best overall performance. To isolate the effect of linguistic rescoring, we further evaluate \textit{BEST}, Transformer, and GRU under identical LM conditions on Brain-to-Text '25 (Table~\ref{tab:lm_controlled}). \textit{BEST} achieves the lowest WER in every setting. Therefore, its advantage is already present before strong language-model priors are introduced.

Notably, weaker encoders benefit more from stronger LMs in absolute terms. Relative to no LM, the current 5-gram reduces WER by 0.0424 for \textit{BEST}, compared with 0.0671 for Transformer and 0.0552 for GRU; the same trend holds for the 4-gram LM. Hence, \textit{BEST}'s gain cannot be attributed to stronger LM utilization. Instead, linguistic rescoring partially compensates for encoder errors without changing the model ranking: under the same 5-gram LM, \textit{BEST} achieves 0.0257 WER, compared with 0.0381 for Transformer and 0.0541 for GRU. Overall, these controlled results indicate that language models reduce residual sequence uncertainty, whereas the dominant performance gain arises from the quality of the neural representation and the resulting decoding space.

\begin{figure*}[!t]
    \centering
    \includegraphics[width=\linewidth]{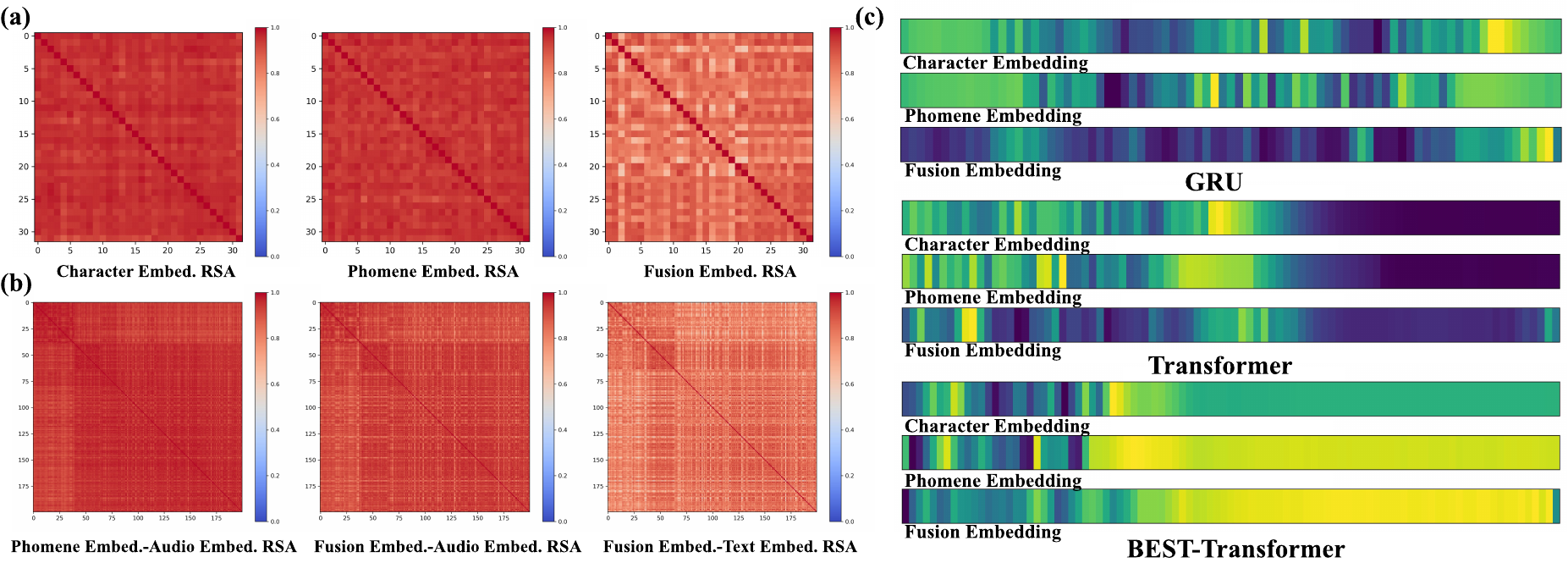}
    \caption{\textbf{(a) Self-RSA on character, phoneme, fusion embedding. (b) Cross-RSA between Phoneme Embedding-Audio Embedding, Phoneme Embedding-Audio Embedding, and Fusion Embedding-Text Embedding. (c) Temporal activation patterns of GRU, transformer, and proposed \textit{BEST}. }}
    \label{fig:visualization1}
\end{figure*}

\textbf{Representation evidence analysis.}
We provide representation-level evidence for the staged decoding mechanism of \textit{BEST} (Fig.~\ref{fig:visualization1}, Appendix \ref{sec:Audio/Text}). Self-RSA reveals that character, phoneme, and fusion embeddings all exhibit clear structure, indicating that neural activity is organized into stable linguistic manifolds rather than diffuse features. Notably, the fusion embedding remains structured but less locally redundant, suggesting a more compact representation after cross-level integration.
Alignment analysis shows that phoneme embeddings weakly match the audio embedding space, whereas fusion embeddings achieve substantially stronger alignment, indicating that the fusion module transforms neural features into an acoustic-compatible manifold suitable for Whisper decoding. This alignment further shifts toward the text embedding space after the Whisper encoder, forming a coherent progression from neural to audio-aligned to text-aligned representations.

Temporal activation patterns support this view: \textit{BEST} produces smoother trajectories with shorter effective support, reflecting stronger temporal consistency and reduced redundancy. In contrast, GRU and vanilla Transformer exhibit fragmented activations, indicating weaker temporal organization.
Overall, these results reveal a concrete mechanism: \textit{BEST} performs decoding via progressive cross-modal alignment. Rather than committing early to discrete phoneme predictions, it first constructs an acoustic-compatible latent space and then maps it to text, providing evidence for a neural-to-audio-to-text decoding pathway.
\begin{table*}[t]
\centering
\caption{\textbf{LM-controlled comparison on Brain-to-Text '25.
All neural encoders are evaluated under identical language-model conditions.
$\Delta$ denotes the absolute WER reduction relative to the corresponding baseline LM setting.}}
\label{tab:lm_controlled}

\normalsize
\renewcommand{\arraystretch}{1.05}
\setlength{\tabcolsep}{2.2pt}

\begin{adjustbox}{max width=\textwidth,center}
\begin{tabular}{lcccccccc}
\toprule

\multirow{2}{*}{\textbf{Method}}
& \multicolumn{4}{c}{\textbf{WER $\downarrow$}}
& \multicolumn{2}{c}{\textbf{$\Delta$ from No LM}}
& \multicolumn{2}{c}{\textbf{$\Delta$ from Mismatched 5-gram}} \\

\cmidrule(lr){2-5}
\cmidrule(lr){6-7}
\cmidrule(lr){8-9}

&
\textbf{No LM}
& \shortstack{\textbf{Mismatched}\\\textbf{5-gram}}
& \shortstack{\textbf{Current}\\\textbf{4-gram}}
& \shortstack{\textbf{Current}\\\textbf{5-gram}}
& $\rightarrow$ \textbf{4-gram}
& $\rightarrow$ \textbf{5-gram}
& $\rightarrow$ \textbf{4-gram}
& $\rightarrow$ \textbf{5-gram} \\

\midrule

\textit{BEST}
& \textbf{0.0681}
& \textbf{0.0624}
& \textbf{0.0292}
& \textbf{0.0257}
& 0.0389
& 0.0424
& 0.0332
& 0.0367 \\

Transformer
& 0.1052
& 0.0875
& 0.0409
& 0.0381
& \textbf{0.0643}
& \textbf{0.0671}
& \textbf{0.0466}
& \textbf{0.0494} \\

GRU
& 0.1093
& 0.0926
& 0.0587
& 0.0541
& 0.0506
& 0.0552
& 0.0339
& 0.0385 \\

\bottomrule
\end{tabular}
\end{adjustbox}

\end{table*}

\begin{figure*}[!t]
    \centering
    \begin{minipage}{0.48\linewidth}
        \centering
        \includegraphics[width=\linewidth]{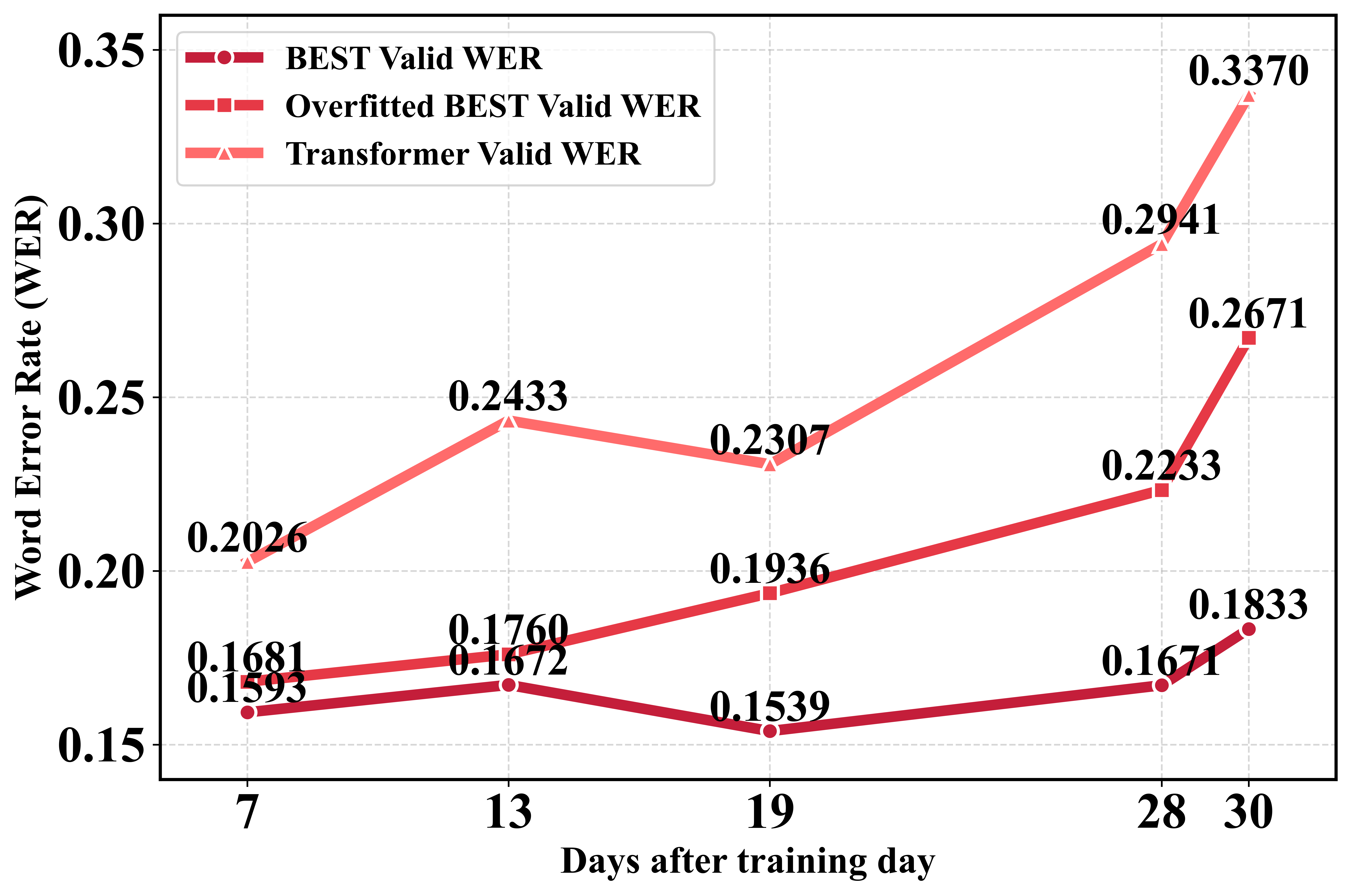}
    \end{minipage}
    \hfill
    \begin{minipage}{0.48\linewidth}
        \centering
        \includegraphics[width=\linewidth]{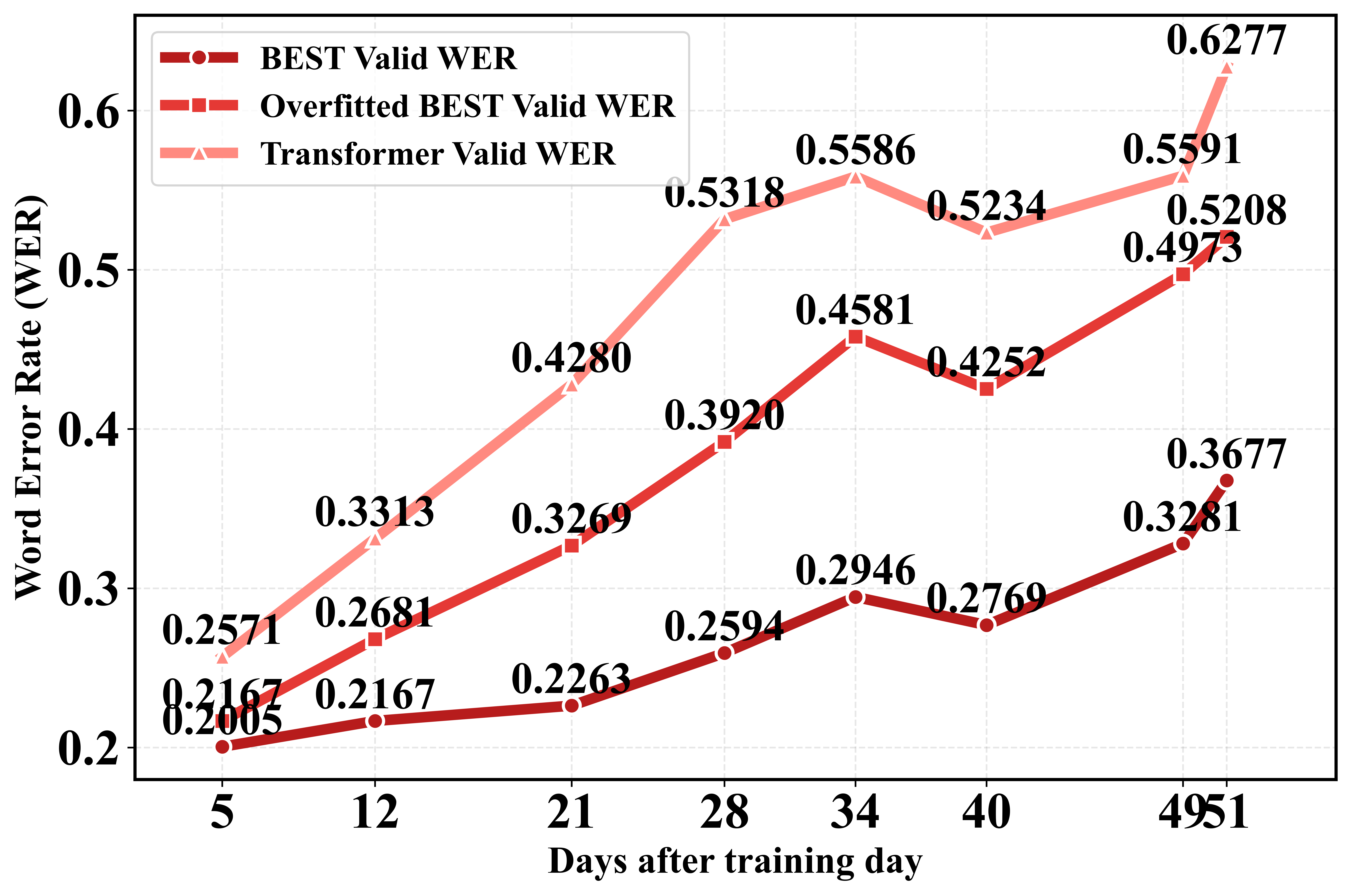}
    \end{minipage}
    \vspace{-2mm}
    \caption{\textbf{Cross-day generalization under temporal shift. Models are trained on first 15 days (left) and 18 days (right), and evaluated on subsequent days.}}
    \label{fig:wer_days}
    \vspace{-2mm}
\end{figure*}
\begin{table*}[!b]
\centering
\vspace{-3mm}
\caption{\textbf{All end-to-end comparable frameworks are reported using inference on Jetson Orin Nano (8GiB) and training on NVIDIA RTX 4090 GPU. Epoch time is reported as mean $\pm$ standard deviation (T12, $\textbf{N=1200}$). GFLOPS stands for floating-point operations per second.}}
\begin{adjustbox}{width=1\columnwidth}
\begin{tabular}{lcccc}
\toprule
\textbf{Model} & \textbf{GFLOPS} & \textbf{Inference times (ms)} & \textbf{Peak Memory (GiB)} & \textbf{Training Epoch time (s)} \\
\midrule 
BIT \cite{zhang2026}    & 1736.26 & 1189.71$\pm$26.81 & 7.23$\pm$0.12 & 147.81$\pm$1.09 \\
\rowcolor{gray!20} \textit{\textbf{BEST (Ours)}}   & \textbf{123.17}  & \textbf{215.67$\pm$19.81}  & \textbf{2.79$\pm$0.06} & \textbf{56.34$\pm$0.81} \\
\bottomrule
\end{tabular}
\end{adjustbox}
\vspace{-5mm}
\label{tab:model_performance}
\end{table*}

\textbf{Cross-day generalization analysis}: We evaluate cross-day generalization by training on the first 15 or 18 days and testing on all subsequent days (Figure~\ref{fig:wer_days}). \textit{BEST} consistently achieves the lowest WER and degrades substantially more slowly than both the overfitted \textit{BEST} and the Transformer baseline as the temporal gap increases. This gap becomes more pronounced in the longer-horizon setting, where both baselines exhibit clear performance drift. These results indicate that the advantage of \textit{BEST} lies in robustness to longitudinal distribution shift rather than improved in-distribution fitting. The overfitted model highlights a key failure mode: stronger fitting leads to faster degradation, revealing sensitivity to day-specific correlations. The Transformer degrades more severely, suggesting that generic sequence modeling is insufficient under cross-session non-stationarity. In contrast, the slower degradation of \textit{BEST} implies that it learns more stable neural-linguistic representations that remain effective across sessions.

\textbf{Deployment analysis of end-to-end inference}: We evaluate deployment efficiency on Jetson Orin Nano (Table \ref{tab:model_performance}). \textit{BEST} reduces computation from 1736.26 to 123.17 GFLOPS, lowers inference latency from 1189.71 ms to 215.67 ms, and cuts peak memory from 7.23 GiB to 2.79 GiB, with training time reduced by over a factor of two. These improvements are substantial, placing the model within the practical constraints of edge deployment. End-to-end decoding is desirable in speech BCI as it enables direct neural-to-text inference without external pipelines, reducing system complexity and latency. However, existing end-to-end frameworks incur prohibitive computational cost. \textit{BEST} retains the advantages of end-to-end decoding while significantly improving efficiency, indicating that the overhead in prior models stems from inefficient representation processing rather than necessary computation. By constructing structured representations before decoding, \textit{BEST} reduces optimization and inference costs, achieving a comparable accuracy-efficiency trade-off for deployment.
\section{Discussion}
\label{sec:Discussion}
\textbf{Rethinking scaling in speech BCI}:
\label{sec:Rethinking}
A central insight of this work is that model scaling does not directly translate to performance gains in speech BCI. Unlike conventional NLP or ASR, speech BCI operates in a low-resource, high-noise regime where neural observations are weakly aligned with linguistic outputs. Under such conditions, increasing model capacity amplifies overfitting rather than improving generalization. Our results show a clear non-monotonic scaling behavior, where compact, modality-aligned decoders outperform larger text-based LLMs. This suggests that the dominant factor in speech BCI is not model size, but the alignment between representation and neural space.

\textbf{Representation, not prediction, is the bottleneck}:
\label{sec:Representation}
We identify the discrete phoneme interface as a key limitation in existing pipelines. Collapsing neural activity into hard phoneme labels discards structured information and leads to brittle decoding. \textit{BEST} avoids this bottleneck by preserving continuous, multi-level representations and performing feature-level integration prior to decoding. Both quantitative and representation-level analyses show that this design improves temporal consistency, reduces representation collapse, and enables more stable sequence reconstruction. This highlights a broader principle: in speech BCI, decoding quality is governed by representation structure rather than intermediate prediction accuracy.

\textbf{Implications for practical deployment}:
\label{sec:Implications}From a systems perspective, \textit{BEST} demonstrates that efficient neural decoding does not require large-scale generative models. By replacing LLM-based decoding with a compact speech-aligned translator, \textit{BEST} significantly reduces computation, latency, and memory footprint while maintaining strong performance. This is critical for clinical deployment, where real-time inference and on-device constraints are fundamental requirements rather than optimization targets.

\textbf{Limitations and future directions}:
\label{sec:Limitations} Despite these advances, several limitations remain. First, our evaluation is conducted on single-subject datasets, and cross-subject generalization remains an open challenge due to variability in neural representations. Second, while \textit{BEST} improves robustness to cross-day distribution shift, long-term non-stationarity and adaptive decoding are not explicitly addressed. Third, our framework relies on pretrained speech models (e.g., Whisper), whose inductive biases may not fully capture the neural-to-linguistic mapping. Future work may explore jointly learned neural-speech representations or fully neural-native decoding architectures.
Finally, while we show that large models are not necessary for current speech BCI settings, this conclusion may evolve with larger-scale neural datasets. Understanding how representation learning and model scaling interact under improved data regimes remains an important direction.

\section{Conclusion}
We present \textit{BEST}, a lightweight and practical framework for speech BCI that replaces discrete phoneme-based pipelines with hierarchical feature-level decoding. \textit{BEST} achieves strong performance with a fraction of the computational cost, while providing improved robustness to overfitting and cross-day distribution shift. Our findings challenge the prevailing assumption that larger models are inherently beneficial for speech BCI. Instead, we show that performance is governed by representation alignment, structured decoding, and controlled capacity. These results suggest a shift in focus from model scaling to representation design, offering a more efficient and robust pathway toward real-world neural speech decoding.
\section*{Acknowledgments}
This work is partly supported by National Natural Science Foundation of China under Grant No. 72471140, partly by Shanghai Jiao Tong University Key Projects of Smart Liberal Arts under Grant No. ZHWK2505, partly by Fundamental Research Funds for the Central Universities under Grant No. YG2025QNA13, partly by Shanghai Magnolia Talent Program-Pujiang Project under Grant No. 2025PJC167, partly by the Lushan Research Funding.
\section*{Conflict of Interest}
The authors declare that they have no competing interests or conflicts of interest related to this work.
\clearpage
\bibliographystyle{plainnat}
\bibliography{example}
\clearpage
\appendix
\section*{Contents}

\noindent
{A. Implementation details}\dotfill \pageref{sec:Implementation details}\\[1.2em]
{\hspace*{2.5em}1 Data preprocessing}\dotfill \pageref{sec:Data preprocessing}\\[1.2em]
{\hspace*{2.5em}2 Training details}\dotfill \pageref{sec:Training details}\\[1.2em]
{B. Audio/Text embedding extraction for RSA}\dotfill \pageref{sec:Audio/Text}\\[1.2em]
{\hspace*{2.5em}1 Audio embedding}\dotfill \pageref{sec:Audio embedding}\\[1.2em]
{\hspace*{2.5em}2 Text embedding}\dotfill \pageref{sec:Text embedding}\\[1.2em]
{\hspace*{2.5em}3 Alignment for RSA}\dotfill \pageref{sec:Alignment for RSA}\\[1.2em]
{C. RNN baseline hyperparameter}\dotfill \pageref{sec:RNN baseline hyperparameter}\\[1.2em]
{D. LLM decoder hyperparameter}\dotfill \pageref{sec:LLM decoder hyperparameter}\\[1.2em]
{E. Phoneme probability visualization}\dotfill \pageref{sec:Phoneme probability visualization}\\[1.2em]
{F. Validation CTC Loss and CER analysis}\dotfill \pageref{sec:Validation CTC loss and CER analysis}\\[1.2em]
{G. Hyperparameter scanning on \textit{BEST} }\dotfill \pageref{sec:Hyperparameter}\\[1.2em]
{\hspace*{2.5em}1 Encoder dimension}\dotfill \pageref{sec:Hyperparameter scanning on encoder dimension}\\[1.2em]
{\hspace*{2.5em}2 Phoneme-character fusion strategy}\dotfill \pageref{sec:Phoneme-Character fusion strategy}\\[1.2em]
{\hspace*{2.5em}3 Temporal masking}\dotfill \pageref{sec:Temporal masking}\\[1.2em]
{\hspace*{2.5em}4 Hierarchical layer allocation}\dotfill \pageref{sec:Hierarchical layer allocation}\\[1.2em]
{\hspace*{2.5em}5 Cascade encoder Last M Layer}\dotfill \pageref{sec:Cascade encoder last M layer}\\[1.2em]
{\hspace*{2.5em}6 Frozen decoder first N Layer}\dotfill \pageref{sec:Froze Decoder first N layer}\\[1.2em]
{H. Phoneme and character confusion analysis}\dotfill \pageref{sec:Phoneme and character confusion analysis}\\[1.2em]
{I. Phoneme and character transition analysis}\dotfill \pageref{sec:Phoneme and character transition analysis}\\[1.2em]
{J. Method implementation}\dotfill \pageref{sec:method}\\[1.2em]
{\hspace*{2.5em}1 Multi-frequency attention}\dotfill \pageref{sec:multi-frequency}\\[1.2em]
{\hspace*{2.5em}2 Hierarchical transformer}\dotfill \pageref{sec:hierarchical}\\[1.2em]
{K. Ethical considerations}\dotfill \pageref{sec:ethical}\\[1.2em]
{L. NeurIPS Paper Checklist}\dotfill \pageref{sec:checklist}\\[1.2em]
\clearpage
\vspace{-3mm}
\section{Implementation details}
\label{sec:Implementation details}
\textbf{Data preprocessing.}
\label{sec:Data preprocessing}
We follow the preprocessing protocols of T12 \cite{Willett2023} and T15 \cite{Card2024Accurate}. 
For \textbf{T12}, 256 neural features are extracted (two features per electrode, including $-4.5$ RMS threshold crossings and spike band power across 128 electrodes), with White Noise Standard Deviation$=0.8$, Gaussian smoothing width$=2.0$, and Constant Offset Standard Deviation$=0.2$. 
For \textbf{T15}, 512 neural features are used (two features per electrode across 256 electrodes), with White Noise Standard Deviation$=1.0$, Gaussian smoothing width$=2.0$, and Constant Offset Standard Deviation$=0.2$.
\begin{table}[t]
\caption{Hyperparameter settings for T12 and T15 across Stage 1 and Stage 2.}
\label{tab:details}
\centering
\begin{adjustbox}{width=0.8\columnwidth}
\begin{tabular}{lcccc}
\toprule
\textbf{Hyperparameter} & \textbf{T12 Stage 1} & \textbf{T12 Stage 2} & \textbf{T15 Stage 1} & \textbf{T15 Stage 2} \\
\midrule
Learning Rate        & $1\times10^{-3}$   & $1\times10^{-5}$ & $1\times10^{-2}$    & $1\times10^{-4}$ \\
Weight Decay         & $1\times10^{-5}$ & $1\times10^{-5}$ & $1\times10^{-3}$   & $1\times10^{-3}$ \\
Batch Size           & 64      & 32      & 64      & 32 \\
Input Dropout        & 0.2     & 0.2     & 0.2     & 0.2 \\
Dropout              & 0.35    & --      & 0.35    & -- \\
Patch Size           & (5, 256)& (5, 256)& (4, 512)& (4, 512) \\
Beam Size           & - & 50 & -& 50 \\

\midrule
Optimizer            & \multicolumn{4}{c}{AdamW} \\
Maximum Sequence     & \multicolumn{4}{c}{500} \\
Embedding Dimension  & \multicolumn{4}{c}{768} \\
Activation           & \multicolumn{4}{c}{ReLU} \\
MLP Dim Ratio        & \multicolumn{4}{c}{4} \\
$\beta_1$            & \multicolumn{4}{c}{0.9} \\
$\beta_2$            & \multicolumn{4}{c}{0.999} \\
\bottomrule
\end{tabular}
\end{adjustbox}
\end{table}

\textbf{Training details:}
\label{sec:Training details}
\textbf{Stage~1} is trained from random initialization on a single NVIDIA RTX~4090 (24\,GiB). Convergence is typically achieved within 80 epochs (under 1 hour for T12 and under 2 hours for T15). Beyond this point, performance gradually degrades due to overfitting, with negligible improvements after 400 epochs as indicated by PER/CER.
\textbf{Stage~2} is trained on a single NVIDIA RTX~4090 (24\,GiB) for 50 epochs (under 1 hour for T12 and under 2 hours for T15). Overfitting is observed beyond 50 epochs, with no significant improvement in WER. The Whisper model is initialized from pretrained weights, with only the first Transformer decoder layer and the fusion block updated during training. Peak system memory usage remains below 32\,GiB. Detailed hyperparameter configurations are provided in Table~\ref{tab:details}.
\vspace{-3mm}
\section{Audio/Text embedding extraction for RSA}
\label{sec:Audio/Text}
\textbf{Audio embedding}:
\label{sec:Audio embedding}
The audio embedding is obtained from the Whisper encoder. We input ground-truth speech audio into the pretrained Whisper encoder and extract the hidden states from the top encoder layer. These representations serve as the reference acoustic embedding space. For consistency with neural-derived features, we use the time-aligned encoder outputs after positional encoding, without additional pooling unless otherwise specified. In experiments involving neural inputs, the fusion embeddings are projected into the same latent space and passed through the upper layers of the Whisper encoder, ensuring that both embeddings lie in a shared acoustic representation.

\textbf{Text embedding}:
\label{sec:Text embedding} The text embedding is obtained from the Whisper decoder. Given the text, we first tokenize the text via Whisper's tokenizer and feed it into the decoder with teacher forcing. We then extract the decoder hidden states corresponding to each token from the final decoder layer. These hidden states are used as the text embedding space. No logits or output projections are used; instead, we rely on the internal decoder representations prior to the vocabulary projection layer.

\textbf{Alignment for RSA}:
\label{sec:Alignment for RSA}For cross-modal RSA, neural-derived embeddings (phoneme, character, and fusion representations) are compared against audio and text embeddings after temporal alignment. We apply linear interpolation or average pooling to match sequence lengths across modalities. All embeddings are normalized before similarity computation to ensure scale invariance.
\vspace{-3mm}
\section{RNN baseline hyperparameter}
\label{sec:RNN baseline hyperparameter}
The hyperparameters for the RNN baseline and LoRA-based LLM decoder are summarized in Table \ref{tab:rnn_params}. For the RNN baseline, we use a regularized training setup with moderate model capacity to ensure stable optimization under limited, noisy neural data, and adopt a warm-up learning rate schedule to enhance early-stage convergence.
\begin{table}[t]
\centering
\caption{Hyperparameters of the RNN Baseline Model}
\label{tab:rnn_params}
\begin{tabular}{lc}
\toprule
\textbf{Hyperparameter} & \textbf{Value} \\
\midrule
GRU Layers / Units & 5 / 768 \\
GRU Dropout & 0.4 \\
Patch Size / Stride & 14 / 4 \\
Input Dropout & 0.2 \\
\midrule
Optimizer & Adam ($\beta_1$=0.9, $\beta_2$=0.999) \\
Weight Decay & $1\times10^{-3}$ \\
LR Warm-up Steps & 1000 \\
Max LR & $5\times10^{-3}$ \\
Min LR & $1\times10^{-4}$ \\
\bottomrule
\end{tabular}
\end{table}

\begin{table}[t]
  \centering
  \vspace{-3mm}
  \caption{Hyperparameter settings for LoRA fine-tuning.}
  \label{tab:lora_hyperparams}
  \begin{adjustbox}{width=0.618\columnwidth}
  \begin{tabular}{lc}
    \hline
    \textbf{Hyperparameter} & \textbf{Value} \\
    \hline
    Activation & ReLU \\
    Learning Rate & $5 \times 10^{-5}$ \\
    Weight Decay & $1 \times 10^{-5}$ \\
    Batch Size & 8 \\
    Gradient Accumulation Step & 1 (model $< 7$B) and 8 (model $\geq 7$B) \\
    LoRA Rank & 8 \\
    LoRA Scaling Factor & 32 \\
    LoRA Dropout Ratio & 0.2 \\
    \hline
  \end{tabular}
  \end{adjustbox}
\end{table}
\vspace{-3mm}
\section{LLM decoder hyperparameter}
\label{sec:LLM decoder hyperparameter}
The hyperparameter of LLM-based decoders \cite{touvron2023llamaopenefficientfoundation, yang2024qwen2technicalreport} is summarized in Table \ref{tab:lora_hyperparams}. we adopt LoRA for parameter-efficient adaptation. The configuration is chosen to provide sufficient adaptation capacity while maintaining training stability and preventing overfitting. Gradient accumulation is adjusted based on model scale to accommodate hardware constraints without affecting optimization dynamics. They follow standard practices and are kept consistent across experiments to ensure fair comparison.
\begin{figure*}[t]
  \centering
  \begin{subfigure}{0.48\linewidth}
    \centering
    \includegraphics[width=\linewidth]{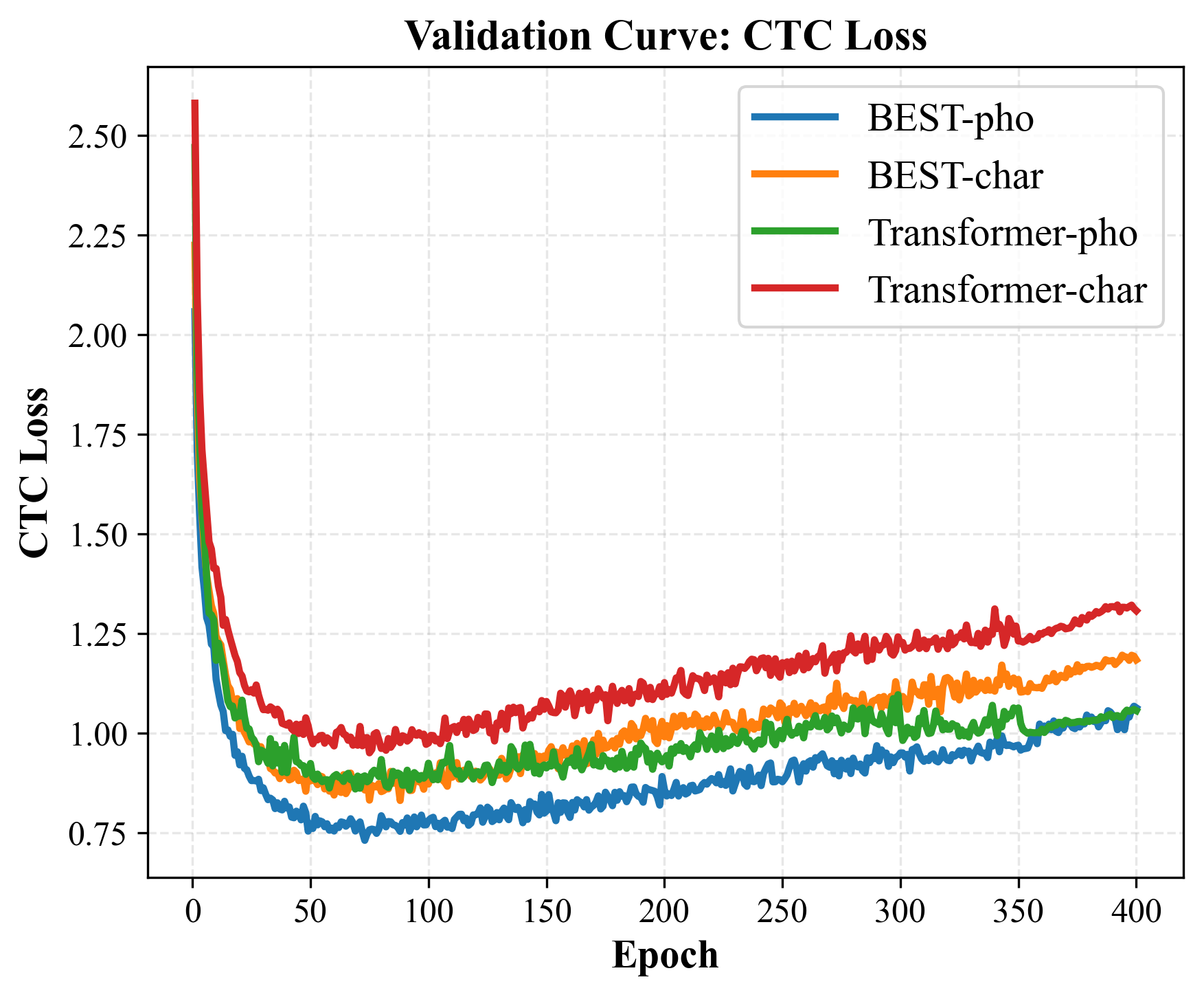}
    \vspace{-3mm}
    \caption{CTC Loss curve}
    \label{fig:loss_curve}
  \end{subfigure}
  \hfill  
  \begin{subfigure}{0.48\linewidth}
    \centering
    \includegraphics[width=\linewidth]{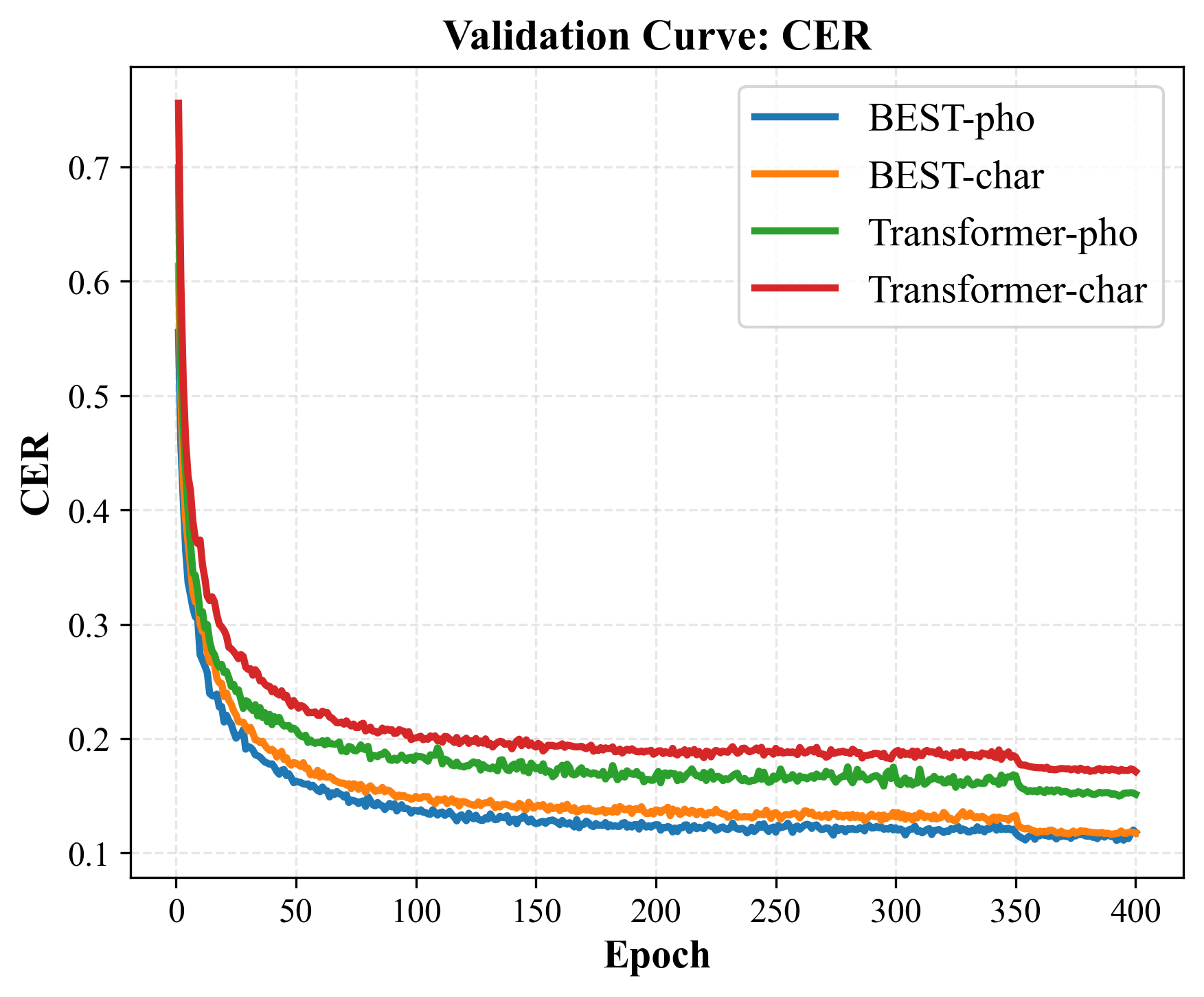}
    \vspace{-3mm}
    \caption{CER curve}
    \label{fig:cer_curve}
  \end{subfigure}
  \vspace{-2mm}
  \caption{Validation curves of CTC loss and CER for the Transformer and proposed \textit{BEST} on participant T12.}
  \vspace{-3mm}
  \label{fig:training_curves}
\end{figure*}
\vspace{-3mm}
\section{Validation CTC loss and CER analysis}
\label{sec:Validation CTC loss and CER analysis}
The validation curves (Figure. \ref{fig:training_curves}) show a consistent optimization advantage of \textit{BEST} over the vanilla Transformer in both phoneme- and character-level pretraining. \textit{BEST} reaches lower CTC loss minima and maintains lower CER throughout training, indicating more stable and more effective neural representation learning. The Transformer, by contrast, converges to a higher loss floor and consistently worse CER, reflecting earlier overfitting and weaker generalization.

A second observation is the divergence between validation CTC loss and CER, which holds across all models. After the early training stage, CTC loss rises while CER continues to decrease or remains stable. This is a property of the CTC objective rather than a model-specific effect. CTC loss depends on the full alignment distribution, whereas CER depends only on the final decoded sequence. A worse alignment distribution therefore does not necessarily imply worse decoding. This highlights a limitation of validation CTC loss as a direct proxy for sequence-level quality.

The advantage of \textit{BEST} remains consistent under both phoneme and character supervision, indicating that the gain is not target-specific. It arises from the encoder design, which improves both local discrimination and sequence consistency. These curves therefore provide optimization-level evidence that structured representation learning, rather than increased model capacity, is the critical factor in speech BCI.
\begin{figure}[t]
    \centering
    \includegraphics[width=0.9\linewidth]{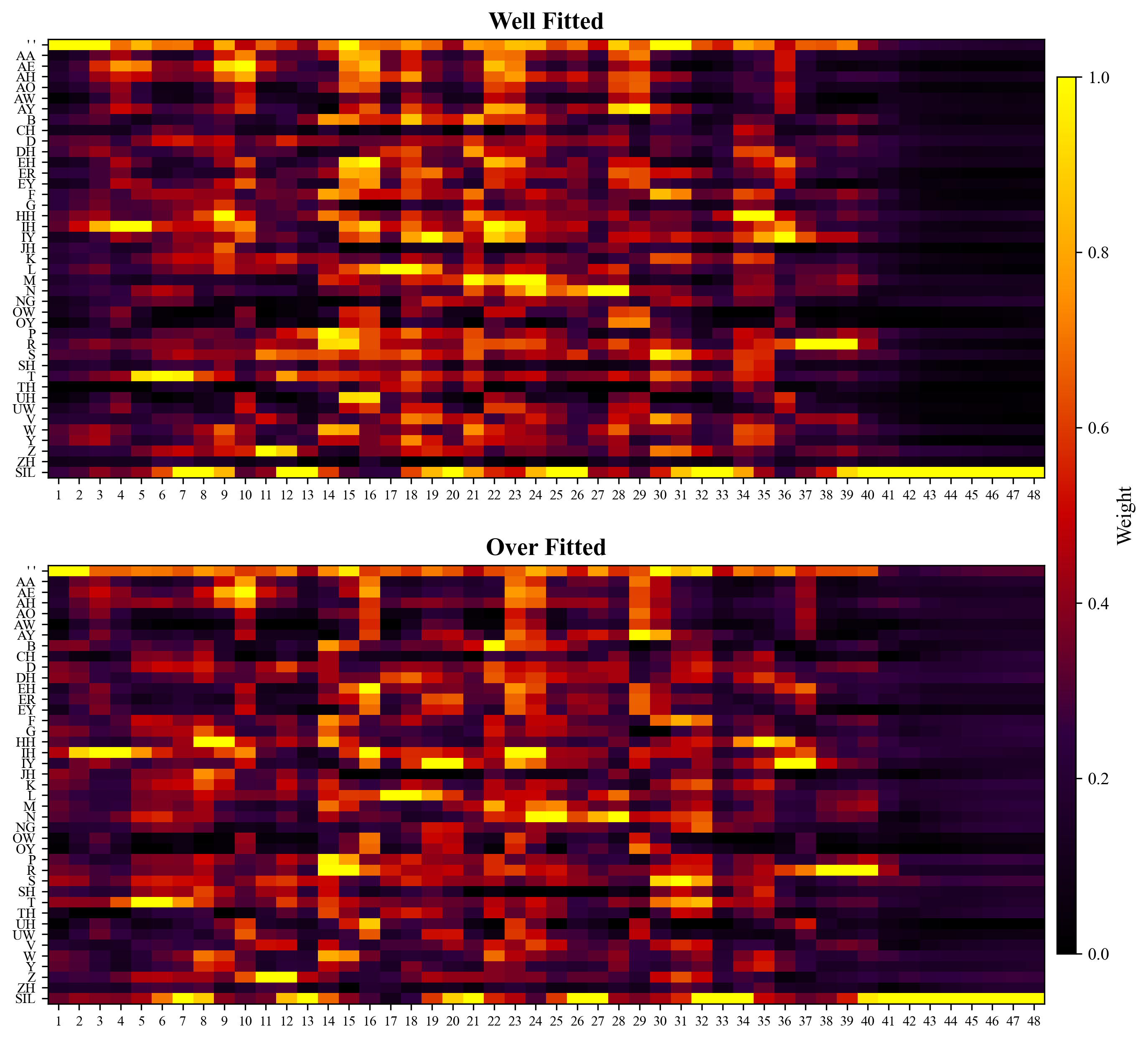}
    \caption{Phoneme decoding probability heatmaps under well-fitted (top) and overfitted (bottom) conditions. Overfitting produces fragmented and overconfident activations, indicating representation collapse despite lower phoneme-level error.}
    \vspace{-3mm}
    \label{fig:phoneme_heatmap}
\end{figure}
\section{Phoneme probability visualization}
\label{sec:Phoneme probability visualization}
Figure \ref{fig:phoneme_heatmap} compares frame-wise phoneme decoding probabilities under well-fitted and overfitted training regimes. The overfitted model produces sharper but more fragmented activations, indicating higher local confidence with weaker temporal organization. By contrast, the well-fitted model maintains smoother transitions and more coherent phoneme trajectories across time. This visualization exposes the mechanism behind the WER degradation of the overfitted model. Overfitting does not primarily destroy local phoneme separability; it disrupts the temporal structure required for consistent sequence decoding. The result is a representation that appears more confident locally but is less compatible with downstream inference. This provides qualitative support for the central claim of the paper: in speech BCI, sequence-level performance depends more on temporal coherence than on isolated phoneme accuracy.
\begin{table}[t]
  \centering
  \caption{Hyperparameter scanning on the dimension of the Transformer encoder. Note that overfitting becomes increasingly prominent during training as the dimension grows.}
  \label{sec:training_ablation}
  \begin{adjustbox}{width=0.618\columnwidth}
  \begin{tabular}{lcc|cc}
    \toprule[1pt]  
    \textbf{dimension} & \textbf{T12 PER$\downarrow$} & \textbf{T12 CER$\downarrow$} & \textbf{T15 PER$\downarrow$} & \textbf{T15 CER$\downarrow$} \\
    \midrule
    384 & 0.1830 & 0.1933 & 0.0971 & 0.1006 \\
    512 & 0.1628 & 0.1708 & 0.0762 & 0.0937 \\
    640& 0.1539 & 0.1627 & 0.0730 & 0.0971 \\
    768 & \textbf{0.1321} & \textbf{0.1436} & \textbf{0.0688} & \textbf{0.0896}\\
    896 & 0.1706 & 0.1772 & 0.0893 & 0.0931 \\
    1024 & 0.2063 & 0.1960& 0.1104 & 0.1052 \\
    \bottomrule[1pt]  
  \end{tabular}
  \end{adjustbox}
  \vspace{-2mm}
\end{table}
\section{Hyperparameter scanning on \textit{\textit{BEST}} configuration}
\label{sec:Hyperparameter}
\textbf{Hyperparameter scanning on encoder dimension}:
\label{sec:Hyperparameter scanning on encoder dimension}We observe a clear non-monotonic dependence on encoder dimension (Table \ref{sec:training_ablation}). Performance improves as the dimension increases from 384 to 768, but degrades beyond this point on both T12 and T15, indicating that larger hidden sizes quickly push the model into an overfitting regime under noisy and data-limited supervision. The optimum at 768 is not incidental: besides providing the best balance between capacity and generalization, it also matches the hidden dimension of the Whisper-based decoder. This dimensional consistency likely reduces projection mismatch between neural representations and the downstream speech embedding space, leading to more effective cross-stage alignment.
\begin{table}[t]
  \centering
  \caption{Ablation on different fusion strategies for Phoneme-Character Fusion.}
  \label{tab:fusion_ablation}
  \begin{adjustbox}{width=0.618\columnwidth}
  \begin{tabular}{lccc}
    \toprule[1pt]
    \textbf{Phon.-Char. Fusion} & \textbf{\#Param.} & \textbf{T12 WER$\downarrow$} & \textbf{T15 WER$\downarrow$}\\
    \midrule
    Linear   & 0.59M        & 0.3215 & 0.1329 \\
    MLP        & 2.36M      & 0.2371 & 0.1135 \\
    Cross Attention & 3.54M & 0.4962 & 0.4219 \\
    Conv1$\times$1  & 0.59M    & 0.1653 & 0.0765 \\
    Conv3$\times$3  & 1.77M   & 0.1560 & 0.0782 \\
    Conv1$\times$1, 3$\times$3 & 2.36M    & \textbf{0.1504}  & \textbf{0.0681} \\
    \bottomrule[1pt]
  \end{tabular}
  \end{adjustbox}
\end{table}

\textbf{Phoneme-Character fusion strategy}:
\label{sec:Phoneme-Character fusion strategy}Table \ref{tab:fusion_ablation} shows that fusion quality is determined by structural bias rather than parameter count alone. Linear and MLP fusion remain limited, while cross-attention performs worst despite its larger capacity, suggesting that heavy global interaction is both inefficient and unstable in this low-resource setting. By contrast, convolutional fusion is consistently more effective. In particular, the multi-scale design achieves the best performance with moderate parameter cost, indicating that it provides the right balance between efficiency and expressiveness. The $1\times1$ branch enables lightweight channel mixing, while the $3\times3$ branch captures short-range temporal context; their combination yields structured feature integration without the excessive flexibility that would encourage overfitting.

\begin{table}[t]
\centering
\small
\setlength{\tabcolsep}{4pt}
\caption{Ablation on masked autoencoder-based reconstruction for the Brain-to-Text ’24 and ’25.}
\label{tab:ablation_mask}
\begin{adjustbox}{width=0.618\columnwidth}
\begin{tabular}{lcc|cc}
\toprule
Ablation & \multicolumn{2}{c}{Brain-to-text '24 T12} & \multicolumn{2}{c}{Brain-to-text '25 T15} \\
       \cmidrule(lr){2-3} \cmidrule(lr){4-5} & PER & CER & PER & CER \\
\midrule
\rowcolor{gray!20} \textbf{\textit{BEST}}      & 0.1321 & 0.1436 & 0.0688 & 0.0896 \\
w. Masked Auto & 0.1420 & 0.1483 & 0.0727 & 0.0915 \\
\bottomrule
\end{tabular}
\end{adjustbox}
\vspace{-2mm}
\end{table}
\begin{table}[t]
\centering
\caption{Hyperparameter scanning on the number of time masking and the number of hierarchical layer configuration in phoneme decoding.}
\label{tab:ablation_time_layer}
\begin{adjustbox}{width=0.816\columnwidth}
\begin{tabular}{cccc}
\toprule[1pt]
\textbf{Number of time masks} & \textbf{CTCLoss$\downarrow$} 
& \textbf{Number of layers} & \textbf{CTCLoss$\downarrow$} \\
\midrule
5  & 0.7238$\pm$0.0081 & 1:2:3 & 0.7562$\pm$0.0048 \\
10 & 0.6870$\pm$0.0064 & 2:2:2 & \textbf{0.7293$\pm$0.0039} \\
\rowcolor{gray!20} 15 & \textbf{0.7293$\pm$0.0039} & 3:2:1 & 0.7503$\pm$0.0040 \\
20 & 0.7519$\pm$0.0071 & 4:1:1 & 0.7761$\pm$0.0052 \\
25 & 0.7636$\pm$0.0049 & 2:1:2 & 0.7438$\pm$0.0063 \\
\bottomrule[1pt]
\end{tabular}
\end{adjustbox}
\end{table}
\begin{table}[t]
\centering
\caption{Hyperparameter scanning on cascaded last layers in the encoder and frozen last layers in the decoder for neural phoneme decoding.}
\label{tab:ablation_encoder_frozen}
\begin{adjustbox}{width=0.816\columnwidth}
\begin{tabular}{cccc}
\toprule[1pt]
\textbf{Encoder Last Layer M} & \textbf{CTCLoss$\downarrow$} 
& \textbf{Frozen FirstLayer N} & \textbf{CTCLoss$\downarrow$} \\
\midrule
1 & 0.8614$\pm$0.0420 & 2  & \textbf{0.7293$\pm$0.0039} \\
2 & 0.8426$\pm$0.0068 & 4  & 0.7415$\pm$0.0033 \\
3 & 0.7682$\pm$0.0052 & 8  & 0.7436$\pm$0.0067 \\
\rowcolor{gray!20} 4 & \textbf{0.7293$\pm$0.0039} & 9  & 0.7823$\pm$0.0060 \\
5 & 0.7554$\pm$0.0106 & 10 & 0.8069$\pm$0.0092 \\
\bottomrule[1pt]
\end{tabular}
\end{adjustbox}
\vspace{-2mm}
\end{table}

\textbf{Masked autoencoder ablation}:
\label{sec:masked}
Table~\ref{tab:ablation_mask} reports the impact of introducing a masked autoencoder (MAE) objective during pretraining. Contrary to prior work (e.g., BIT~\cite{zhang2026}), incorporating masking leads to a consistent degradation across both benchmarks, with PER increasing from 0.1321 to 0.1420 (+7.5\%) on T12 and from 0.0688 to 0.0727 (+5.7\%) on T15. A similar trend is observed for CER.
This result suggests a fundamental mismatch between MAE-style reconstruction and the requirements of speech BCI decoding. Specifically, MAE encourages pointwise signal recovery under heavy corruption, which biases the model toward low-level local structure reconstruction. However, neural speech signals are inherently noisy, temporally diffuse, and only indirectly aligned with acoustic realizations. As a result, emphasizing reconstruction can divert model capacity away from learning invariant, linguistically meaningful representations. In particular, phonetic identity and word-level semantics are encoded in distributed temporal-spectral patterns rather than in precisely reconstructable signal segments.
Furthermore, masking disrupts the already sparse and information-limited neural inputs, exacerbating ambiguity in the latent space. This is especially detrimental in low-SNR regimes typical of intracranial or non-invasive recordings, where preserving global temporal coherence is more critical than recovering masked fragments. The relatively larger degradation on T12 also indicates that MAE objectives are less robust under more challenging decoding conditions.

\textbf{Temporal masking}:
\label{sec:Temporal masking} Performance improves when masking is increased from a sparse setting, but degrades once masking becomes too aggressive (Table \ref{tab:ablation_time_layer}). This pattern suggests that temporal masking is beneficial only when it regularizes local redundancy without destroying the continuity of speech-relevant dynamics. Too little masking fails to suppress over-reliance on local correlations, whereas too much masking removes the temporal evidence required for stable phoneme alignment. The optimum therefore lies in a moderate regime, where masking improves invariance while preserving sequence structure.

\textbf{Hierarchical layer allocation}:
\label{sec:Hierarchical layer allocation}As shown in Table~\ref{tab:ablation_time_layer}, the symmetric $2:2:2$ configuration consistently achieves the lowest CTC loss, while increasingly imbalanced allocations lead to noticeable degradation. This suggests that phoneme decoding from neural signals critically depends on a balanced distribution of representational capacity across hierarchical stages. Allocating excessive depth to early layers encourages overfitting to local temporal fluctuations and low-level neural dynamics, limiting the emergence of stable phonetic abstractions. In contrast, concentrating depth in later layers weakens early-stage temporal grounding and delays the progressive aggregation of contextual information. The symmetric design mitigates both issues by enforcing a more uniform transformation pipeline, enabling a gradual transition from fine-grained temporal encoding to higher-level phonetic and contextual representations. This balanced hierarchy is better aligned with the inherently multi-scale and distributed nature of neural speech signals, leading to improved generalization.

\textbf{Cascade encoder last M layer}:
\label{sec:Cascade encoder last M layer}Performance is maximized in an intermediate regime (Table \ref{tab:ablation_encoder_frozen}). Cascading too few encoder layers leads to weak adaptation, as the neural representation remains insufficiently aligned with the downstream decoder space. Cascading too many layers, however, begins to erode the stability of the pretrained structure and introduces unnecessary flexibility, which is harmful under limited supervision. The optimum at 
$M=4$ indicates that effective neural decoding requires substantial but controlled representational transformation before sentence-level decoding.
\begin{figure*}[t]
    \centering \includegraphics[width=0.95\linewidth]{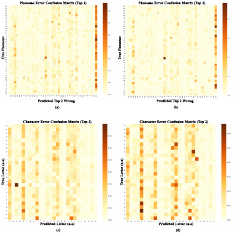}
    \vspace{-3mm}
    \caption{Top-2 phoneme and character confusion matrices on Brain-to-Text '24 and '25 benchmarks.}
    \vspace{-3mm}
    \label{fig:frame45}
\end{figure*}
\begin{figure*}[t]
    \centering \includegraphics[width=0.95\linewidth]{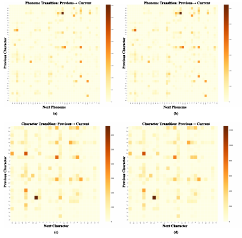}
    \vspace{-3mm}
    \caption{Phoneme and character transition matrices (previous $\rightarrow$ current) on Brain-to-Text '24 and '25 benchmarks. The matrices capture empirical transition patterns in neural decoding at both phonological and orthographic levels.}
    \vspace{-5mm}
    \label{fig:frame5}
\end{figure*}

\textbf{Froze Decoder first N layer}:
\label{sec:Froze Decoder first N layer}The decoder freezing results are consistent with this interpretation (Table \ref{tab:ablation_encoder_frozen}). Freezing only a small number of initial decoder layers yields the best performance, while freezing more layers degrades CTC loss. This suggests that the decoder must retain enough fixed pretrained structure to stabilize optimization, but still leave sufficient trainable capacity near the interface for adaptation to neural-derived inputs. Excessive freezing restricts this interface and prevents effective alignment; insufficient freezing would instead expose too much of the decoder to noisy supervision. Together, these results support a broader conclusion of the paper: in speech BCI, the best operating point is not obtained by maximizing flexibility, but by carefully controlling where adaptation is allowed and where pretrained structure is preserved.
\section{Phoneme and character confusion analysis}
\label{sec:Phoneme and character confusion analysis}
We examine the top-2 phoneme and character confusion matrices on the Brain-to-Text ’24 and ’25 benchmarks (Figure \ref{fig:frame45}). The patterns reveal strong structural regularity: phoneme confusions cluster within articulatorily and acoustically similar groups, while character confusions, though broader due to phoneme-to-grapheme mapping, remain highly organized. These results indicate that neural decoding preserves linguistic structure rather than producing arbitrary errors.

The consistency of confusion patterns across both datasets demonstrates that these error structures are intrinsic to neural representations rather than dataset-specific artifacts. Notably, the dominant phoneme confusions align with natural phonotactic constraints, and character confusions reflect coherent orthographic patterns, highlighting a systematic propagation of uncertainty from phoneme to text space.
Critically, these observations expose a fundamental limitation of discrete phoneme-based decoding. While local predictions are largely plausible, the model accumulates errors across sequences because early commitment to discrete units discards continuous evidence embedded in neural representations. This provides mechanistic support for \textit{BEST}’s design: delaying discrete decisions and integrating multi-level representations preserves structured neural-linguistic information, enabling more reliable sequence-level decoding.
\section{Phoneme and character transition analysis}
\label{sec:Phoneme and character transition analysis}
We analyze phoneme and character transition statistics (previous $\rightarrow$ current) on Brain-to-Text ’24 and ’25 (Figure \ref{fig:frame5} and observe highly structured and sparse patterns. Transitions are concentrated on a small subset of dominant pairs, indicating that neural decoding preserves strong local sequential dependencies rather than producing independent predictions.

At the phoneme level, transitions are largely confined within phonotactically plausible neighborhoods, reflecting preserved articulatory continuity. At the character level, the patterns are more diffuse due to phoneme-to-grapheme ambiguity, yet remain strongly structured. Notably, the transition distributions are highly consistent across both datasets, despite differences in data composition and recording conditions, suggesting that these patterns reflect intrinsic properties of neural speech representations rather than dataset-specific effects.

These results reveal a fundamental limitation: neural decoding captures local continuity but lacks sufficient capacity to resolve ambiguity over longer contexts. Consequently, while local transitions are largely correct, errors accumulate at the sequence level. This further supports the design of \textit{BEST}, which mitigates this limitation by delaying discrete decisions and leveraging richer contextual representations for decoding.
\vspace{-2mm}
\section{Method implementation}
\label{sec:method}
\textbf{Multi-Frequency Attention}:
\label{sec:multi-frequency}
We introduce a lightweight multi-frequency channel attention module to capture informative temporal-spectral patterns in neural signals. Given input $x \in \mathbb{R}^{B \times C \times T}$, we first align the temporal dimension to a fixed length via adaptive average pooling. The signal is then projected onto a set of fixed discrete cosine transform (DCT) bases to extract frequency-specific responses. Specifically, we employ \textbf{16 fixed low-frequency DCT branches} corresponding to indices \textbf{1, 2, 4, 8, 16, 32}, which are selected by the top-$K$ strategy. For each frequency branch, we compute three complementary channel-wise statistics: mean, max, and min. We aggregate these statistics and feed them into a shared bottleneck MLP to produce attention weights.The fixed DCT bases are stored as non-trainable buffers, and the input is reweighted by the sigmoid-activated attention to form final output.This design imposes negligible overhead while boosting robustness to noise and temporal distribution shifts in neural signals.

\noindent\textbf{Pseudo-code:}
\[
\begin{array}{l}
\text{Input: } x \in \mathbb{R}^{B \times C \times T} \\
x_{\text{pool}} = \text{AdaptiveAvgPool}(x, L) \\
\text{feat} = 0 \\
\text{for each DCT basis } f_i: \\
\quad s = x_{\text{pool}} * f_i \\
\quad \text{feat} += \text{Mean}(s) + \text{Max}(s) + \text{Min}(s) \\
\text{feat} = \text{feat} / K \\
\text{attn} = \text{Sigmoid}(\text{MLP}(\text{feat})) \\
\text{Output: } x * \text{attn.unsqueeze}(-1)
\end{array}
\]

\textbf{Hierarchical Transformer}:
\label{sec:hierarchical}
To capture temporal dependencies at multiple scales, we employ a hierarchical Transformer with progressive masking. A shallow initial Transformer layer models global context, followed by stacked Transformer blocks with distinct temporal constraints, including causal (past), local-window, and forward (future) attention. This staged design separates long- and short-range temporal dependencies, improving stability and robustness under noisy neural inputs.

\noindent\textbf{Pseudo-code:}
\[
\begin{array}{l}
\text{Input: } x \\
h = \text{iTransformer}(x) \\
h = \text{Transformer1}(h, \text{mask=past\_mask}) \\
h = \text{Transformer2}(h, \text{mask=local\_mask}) \\
h = \text{Transformer3}(h, \text{mask=future\_mask}) \\
\text{Output: } \text{Projection}(h)
\end{array}
\]
\section{Ethical considerations}
\label{sec:ethical} This work presents a lightweight speech brain-computer interface (BCI) for neural-to-text decoding, with careful consideration of privacy, ethics, and responsible deployment.
All datasets (Brain-to-Text ’24/’25) are publicly available and collected under IRB approval with informed consent. Data are fully anonymized, and no new human experiments are conducted.
Speech BCIs raise risks of neural privacy leakage and unintended cognitive inference. The proposed BEST framework mitigates these risks via a compact architecture that avoids reliance on large third-party models, reducing external data transmission. It is explicitly designed to decode only speech-related neural features, adhering to minimal and task-specific information use.

Clinically, this work targets communication restoration for patients with severe motor impairments. Its lightweight design enables edge deployment, promoting accessibility. However, further clinical validation, safety standardization, and long-term evaluation are required before real-world use.
We further advocate efficient model design over unnecessary scaling, reducing computational cost and environmental impact. Limitations, including single-subject evaluation and limited cross-subject generalization, are clearly acknowledged to avoid overstatement.
Future work will follow medical ethics standards and incorporate secure data governance (e.g., encryption and access control), alongside interdisciplinary collaboration to ensure BCI deployment aligns with privacy, and human dignity.
\section*{NeurIPS Paper Checklist}
\label{sec:checklist}
\begin{enumerate}

\item \textbf{Claims}
\textbf{Answer: Yes.} \\
The claims in the abstract and introduction accurately reflect the contributions and experimental results presented in the paper. All stated assumptions and limitations are consistent with the scope of the work.

\item \textbf{Limitations}
\textbf{Answer: Yes.} \\
The paper includes a discussion of limitations, including assumptions, dataset scope, and potential failure cases, and analyzes their impact on generalization.

\item \textbf{Theory, Assumptions and Proofs}
\textbf{Answer: N/A.} \\
The paper does not include theoretical results or formal proofs.

\item \textbf{Experimental Result Reproducibility}
\textbf{Answer: Yes.} \\
We provide sufficient details of the model architecture, training procedure, and implementation to enable reproducibility of the results.

\item \textbf{Open Access to Data and Code}
\textbf{Answer: Yes.} \\
We provide code, data processing scripts, and instructions necessary to reproduce the main experimental results.

\item \textbf{Experimental Setting / Details}
\textbf{Answer: Yes.} \\
The paper specifies all key experimental details, including dataset splits, hyperparameters, and training configurations.

\item \textbf{Experiment Statistical Significance}
\textbf{Answer: Yes.} \\
We report results averaged over multiple runs with appropriate measures of variability (e.g., standard deviation).

\item \textbf{Experiments Compute Resource}
\textbf{Answer: Yes.} \\
The paper reports the computational resources used, including hardware specifications and training time.

\item \textbf{Code of Ethics}
\textbf{Answer: Yes.} \\
We confirm that the research complies with the NeurIPS Code of Ethics.

\item \textbf{Broader Impacts}
\textbf{Answer: Yes.} \\
The paper discusses potential broader impacts, including both beneficial applications and possible negative societal implications.

\item \textbf{Safeguards}
\textbf{Answer: N/A.} \\
The proposed method does not involve high-risk models requiring specific safeguards for deployment.

\item \textbf{Licenses}
\textbf{Answer: Yes.} \\
We properly cite all datasets, codebases, and pretrained models used in this work and respect their respective licenses.

\item \textbf{Assets}
\textbf{Answer: Yes.} \\
We release new assets associated with this work and provide documentation describing their usage, training details, and limitations.

\item \textbf{Crowdsourcing and Research with Human Subjects}
\textbf{Answer: Yes.} \\
We include details of instructions provided to participants and relevant experimental setup information. Compensation details are also described where applicable.

\item \textbf{IRB Approvals}
\textbf{Answer: Yes.} \\
The study involving human subjects was reviewed and approved by the appropriate institutional review board (IRB) or equivalent ethics committee.

\item \textbf{Declaration of LLM Usage}
\textbf{Answer: Yes.} \\
Large Language Models were used only for writing assistance and did not influence the core methodology or experimental results.

\end{enumerate}
\end{document}